\documentclass[10pt,twocolumn,letterpaper]{article}

\usepackage[algorithms]{wacv}      

\usepackage{graphicx}
\usepackage{amsmath}
\usepackage{amssymb}
\usepackage{booktabs}
\usepackage{diagbox}
\usepackage{multirow}
\usepackage{subcaption}

\usepackage[pagebackref,breaklinks,colorlinks]{hyperref}

\usepackage[capitalize]{cleveref}
\crefname{section}{Sec.}{Secs.}
\Crefname{section}{Section}{Sections}
\Crefname{table}{Table}{Tables}
\crefname{table}{Tab.}{Tabs.}

\def\wacvPaperID{1323} 
\def\confName{WACV}
\def\confYear{2025}

\begin{document}

\title{Relational Self-supervised Distillation with Compact Descriptors \\ for Image Copy Detection}


\author{
    Juntae Kim\textsuperscript{1,2}\thanks{These authors contributed equally to this work.} \quad
    Sungwon Woo\textsuperscript{1}\footnotemark[1] \quad
    Jongho Nang\textsuperscript{1}\thanks{Corresponding author.} \\[0.5em]
    \textsuperscript{1}Sogang University \hspace{2em} \textsuperscript{2}mAy-I \\
    {\tt\small \{jtkim1211, swwoo, jhnang\}@sogang.ac.kr}
}

\maketitle

\begin{abstract}

Image copy detection is the task of detecting edited copies of any image within a reference database. While previous approaches have shown remarkable progress, the large size of their networks and descriptors remains a disadvantage, complicating their practical application. In this paper, we propose a novel method that achieves competitive performance by using a lightweight network and compact descriptors. By utilizing relational self-supervised distillation to transfer knowledge from a large network to a small network, we enable the training of lightweight networks with smaller descriptor sizes. We introduce relational self-supervised distillation for flexible representation in a smaller feature space and apply contrastive learning with a hard negative loss to prevent dimensional collapse. For the DISC2021 benchmark, ResNet-50 and EfficientNet-B0 are used as the teacher and student models, respectively, with micro average precision improving by 5.0\%/4.9\%/5.9\% for 64/128/256 descriptor sizes compared to the baseline method. The code is available at \href{https://github.com/juntae9926/RDCD}{https://github.com/juntae9926/RDCD}.
\end{abstract}

\begin{figure}[t]
\centering
\includegraphics[width=0.9\linewidth]{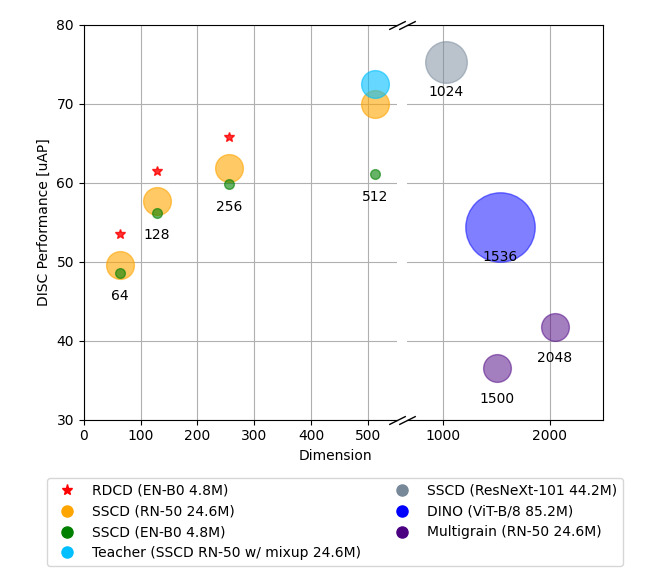}
\caption{Comparison of RDCD(Ours) and other image copy detection methods. RDCD utilizes a lightweight network and achieves high performance with compact descriptor sizes. }
\label{fig:intro}
\end{figure}

\section{Introduction}
\label{sec:intro}

Image copy detection (ICD) is the task of identifying edited copies of images within an existing database. This is widely used in online sharing platforms and social networks to filter content for copyright protection. Over the past years, ICD has been considered a sub-task of instance-level image retrieval~\cite{disc}. These approaches utilize global descriptors~\cite{babenko2014neural, razavian2016visual, radenovic2018fine, mohedano2016bags} and local descriptors~\cite{noh2017large, simeoni2019local, tolias2020learning}, focusing on identifying specific objects that represent the images. However, ICD addresses a more complex challenge by focusing on detecting exact copies of the images themselves, rather than identifying representative objects. It deals with precise duplicates of images that undergo severe transformations such as re-encoding, resizing, merging, cropping, warping, or color distortion. To better address these challenges, recent studies have leveraged self-supervised learning (SSL) methods~\cite{dino, simclr, moco, barlow} to directly apply data augmentation as their training objective~\cite{sscd,wang2021bag}.

ICD involves two key challenges. The first challenge is the large scale of the detection system. In large-scale systems, millions of images in the database are pre-processed offline into descriptors and stored. Online query images are then converted into descriptors in real-time, after which a nearest-neighbor search method is employed, similar to image search systems~\cite{liu2007clustering,guo2020accelerating,johnson2019billion}. However, this large-scale approach often leads to significant computational overhead, resulting in prolonged processing times, especially when dealing with vast image databases. To efficiently improve large-scale systems, an ideal approach is to use smaller architectures or utilize smaller-sized descriptors. This approach effectively reduces search time and storage space while maintaining system performance. However, several studies~\cite{shi2022efficacy,gu2021simple} have found that lightweight networks with limited representation power cannot directly employ SSL to produce acceptable performance. Another method uses principal component analysis (PCA) to create small-sized descriptors, but it leads to significant performance degradation~\cite{sscd}. The second major challenge is the complex task of distinguishing between genuine copies and visually similar but unrelated images, known as hard negative samples. These hard negative samples are prevalent in ICD scenarios and can significantly complicate the identification process. Examples include images captured from different camera angles or images captured at the same location at different times, which are visually similar to the original but are not actual edited copies. A uniform distribution helps to maximize the separation between each image and its closest neighbors, including hard negatives. SSCD~\cite{sscd} employs the Kozachenko-Leononenko estimator~\cite{koleo} as an entropy regularizer to promote this uniform distribution. This technique ensures the distances between different regions in the embedding space are more consistent and comparable, thereby improving the system’s ability to differentiate between genuine copies and hard negative samples. By achieving uniform distribution, this approach makes full use of all dimensions, thereby addressing the issue of dimensional collapse. Despite these advancements, challenges still remain in efficiently training lightweight networks for ICD, especially when dealing with hard negatives, while maintaining high performance.


In this work, we introduce Relational Self-supervised Distillation with Compact Descriptors (RDCD), a novel approach that combines relation-based self-supervised distillation (RSD) and Hard Negative (HN) loss to address these challenges. Our method aims to effectively capture the intricate relationships between teacher model descriptors and represent them in a reduced feature space. Existing SSL with knowledge distillation (SSL-KD) approaches use feature-based knowledge distillation (FKD)~\cite{disco} or RSD~\cite{seed, compress} for pre-training in classification tasks. FKD transfers knowledge by aligning embeddings of the student model with those of the teacher model using mean squared error loss. However, we found that FKD fails to prevent dimensional collapse, leading to performance degradation in ICD. To address this issue, we create a separate instance queue for the teacher to make the student mimic the pairwise similarity generated by the teacher using Kullback-Leibler (KL) divergence, allowing the student network to generate discriminative descriptors in its own feature space.


Although our distillation method is similar to previous SSL-KD methods~\cite{seed, compress}, it ultimately differs in the type of information it aims to distill. The main purpose of these methods is to provide advantages for downstream tasks as pre-training methods in lightweight networks. Their objective is to enrich the intermediate representations with transformation-covariant features, embedding useful information, such as color and orientation~\cite{simclr}. In contrast, our objective is to learn transformation-invariant representations at the final layer rather than obtaining transformation-covariant intermediate features. This design is crucial as the ICD task requires representations that go beyond instance discrimination to a tighter level, which is distinctly different from instance retrieval or classification tasks. In the ICD task, transformation-invariant features are essential, while intermediate representations with transformation-covariant characteristics degrade performance, as illustrated in~\cite{sscd}.



We conduct comparative experiments with SimCLR~\cite{simclr}, MoCo-v2~\cite{MoCov2}, FKD, and RSD to analyze the effectiveness of each method in training efficient lightweight networks for ICD. We test our approach on various teacher and student networks, including convolutional neural networks and vision transformers, to demonstrate its architecture-agnostic characteristics. The main contributions of this work are summarized as follows:

\begin{itemize}
    \item We introduce RDCD, a novel approach that combines RSD and HN loss to train lightweight networks for ICD in a self-supervised manner. Our method leverages the relational information between descriptors generated by a teacher network to guide the learning of a student network with a compact descriptor size.
    \item We employ HN loss to prevent the dimensional collapse that can occur in SSL using compact descriptors, ensuring a structured and informative feature space.
    \item Through extensive experiments, we validate the effectiveness of our RDCD approach with various architectures, including CNNs and ViTs.
\end{itemize}

\begin{figure*}[t]
\centering
\includegraphics[width=0.9\linewidth]{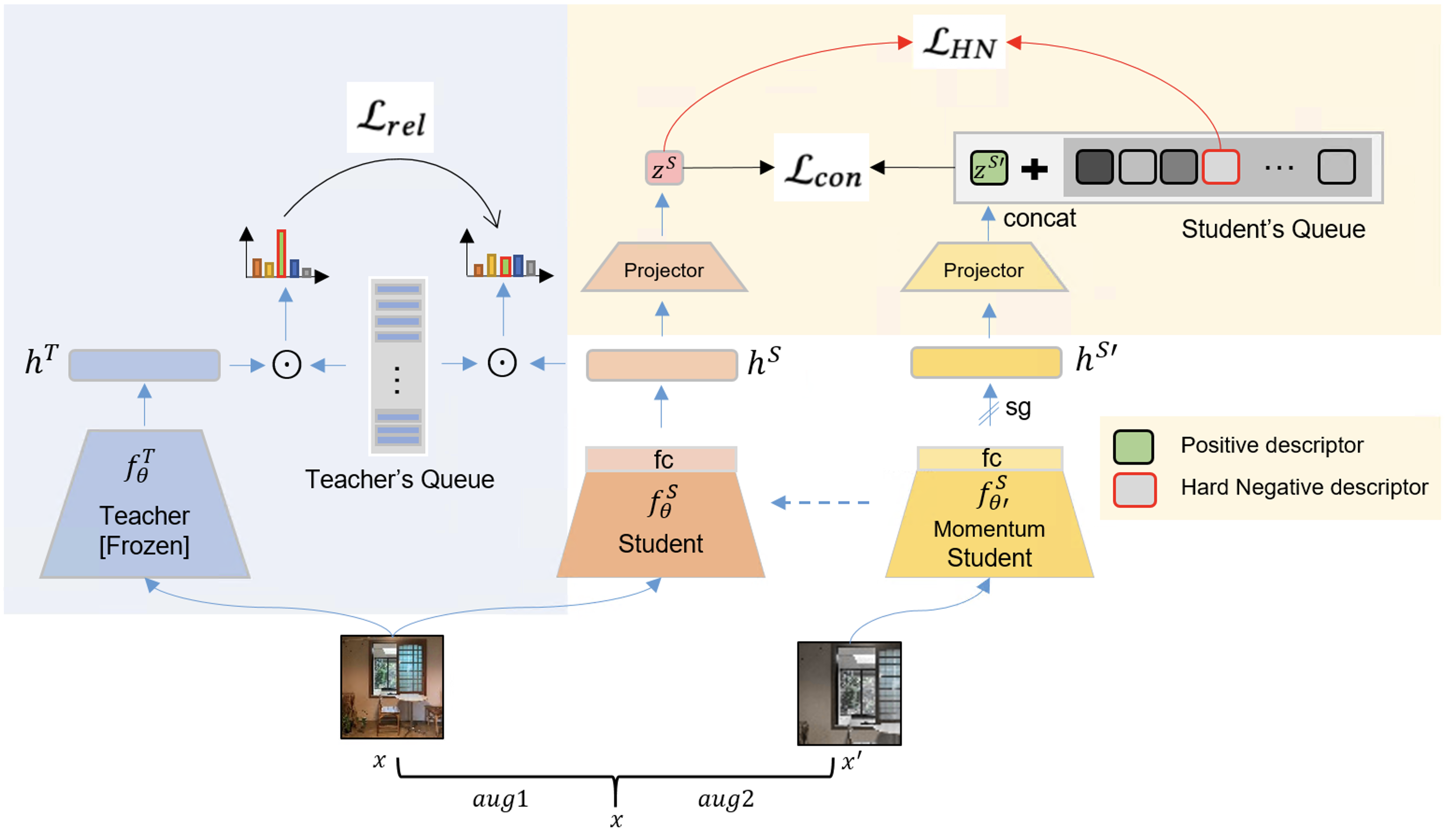}
\caption{The overall pipeline of proposed Relational Self-supervised Distillation for Image Copy Detection (RDCD). The method combines three key components: (1) Relational Self-supervised Distillation (RSD) which transfers knowledge from a pre-trained teacher network to a lightweight student network, (2) Contrastive Learning which uses MoCo-v2, and (3) Hard Negative (HN) Loss to address challenging negative samples. The student network $f^S_\theta$ extracts features $h^S$ and $h^{S'}$ from two augmented views of an input image. These features are then projected to a lower dimension for use in contrastive learning. The teacher network $f^T_\theta$ guides RSD by comparing similarities between instances in its feature space. The final RDCD loss is a weighted combination of RSD loss ($\mathcal{L}_{rel}$), contrastive loss ($\mathcal{L}_{con}$), and HN loss ($\mathcal{L}_{hn}$), enabling the student to learn compact yet effective descriptors for image copy detection.}
\label{fig:architecture}
\end{figure*}

\section{Related Work}
\subsection{Existing Image Copy Detection Methods}
The ICD task has seen significant advancements since the 2021 Image Similarity Challenge~\cite{douze20212021} hosted by Meta AI Research, which spurred numerous research developments. Previous approaches utilize either contrastive self-supervised learning or deep metric learning. CNNCL~\cite{cnncl} uses contrastive loss and cross-batch memory. EFNets~\cite{efnet} employs Arcface loss-based metric learning and uses a drip training technique to incrementally increase the number of classes during training. BOT~\cite{bot} uses deep metric learning through hard sample mining with triplet loss and classification loss. In subsequent years, SSCD~\cite{sscd} applies InfoNCE loss with entropy regularization. Different from previous winning solutions which use numerous techniques to enhance performance, such as ensembles and post-processing, our approach follows the same evaluation dataset and basic evaluation methods from SSCD~\cite{sscd}. Additionally, we trained our model using a purely unlabeled dataset, without employing any ground truth labels. Active Image Indexing~\cite{active} applies imperceptible modifications to images called "activation", when using an ICD model in conjunction with Approximate Nearest Neighborhood indexing, ensuring the transformed image is positioned at the center of the quantized cluster. To address efficiency, our method aims to directly reduce descriptor sizes without relying on additional indexing techniques~\cite{invertedfile, product10quantization}.

\subsection{Knowledge Distillation}
Using a compact model to approximate a function learned by a more comprehensive, higher-performing model was first introduced in~\cite{bucilu2006model}. This concept was expanded in~\cite{hinton2015distilling}, where the student model is trained to mimic the teacher's softened logits, a process known as knowledge distillation. Feature distillation has been garnering significant attention recently. FitNet~\cite{romero2014fitnets} adds a new dimension by proposing a hint-based training scheme for the alignment of feature maps, known as feature distillation, selecting the L2 distance as the metric for comparing the two feature maps. Similarly, \cite{zagoruyko2016paying} suggests the transfer of spatial attention maps from a high-performing teacher network to a smaller student network. \cite{yim2017gift} introduces the idea of transferring 'flow'—defined as the inner product between features from two layers— to the student. PKT~\cite{passalis2018learning} directly aligns the probability distributions of the data between the teacher's and student's feature spaces. The success of these knowledge distillation methods can primarily be attributed to the insightful knowledge embedded in the logits of the teacher model.

\subsection{Relation-based Knowledge Distillation}
While conventional knowledge distillation only extracts the information for a single data point from the teacher, similarity-based distillation methods~\cite{tung2019similarity, compress, seed, passalis2018learning, park2019relational, tian2019contrastive, tejankar2021isd} learn the knowledge from the teacher in terms of similarities between data points. A pairwise similarity matrix has been proposed to retain the interrelationships of similar samples between the representation space of the teacher and student models. CompRess~\cite{compress} and SEED~\cite{seed}, which are approaches related to our method, use similarity-based distillation to compress a large self-supervised model into a smaller one. However, our method differs in that we distill the relationship at the final layer to learn the transformation-invariant descriptor itself, whereas previous approaches enrich the intermediate representations with more transformation-covariant features.


\section{Methodology}
\subsection{Overall Architecture}
The overall architecture of the proposed RDCD approach is presented in Fig~\ref{fig:architecture}. In RDCD, the student network is trained using three different objectives. Because the student network $f^S_\theta$ is the encoder that we want to improve, we freeze teacher network $f^T_\theta$ which is a pre-trained encoder by training using an off-the-shelf method.

Given an image x, the teacher network extracts its representation $h^T$, in which $h^T \in \mathbb{R}^{D_T}$ is the final feature without any classifier. The student network extracts two representations $h^S$ and $h^{S'}$ from input images augmented in different ways. An FC layer is employed immediately after the student network to match the dimensionalities of $h^T$ and $h^S$. Furthermore, we also add an additional projector for contrastive learning by the student. This projector decreases the dimensions of representation and is used to calculate pairwise similarities for the computation of the contrastive loss function. We use SimCLR and MoCo-v2 as our contrastive learning methods, but any contrastive-based learning method can be adopted.

\subsection{Relational Self-supervised Distillation}
Relational Self-supervised Distillation (RSD) is a method that transfers instance relations from a teacher network to a student network. This process involves the creation of an instance queue within the teacher's network, designed to store the teacher's instance embeddings. This queue serves as a reference point for the student network, facilitating the transfer of knowledge. When a new sample is introduced, its similarity scores are calculated against all instances in the queue ($q_k$) using both the teacher and student networks. The similarity scores are typically computed using cosine similarity.

The key objective is to align the similarity score distribution generated by the student network with that of the teacher network. This alignment is achieved by minimizing the KL-divergence between the similarity score distributions of the two networks, ensuring a close match in their respective interpretations of the instance similarities. We apply the same augmentations into an image $x_i$ as a batch and map the embeddings $h^T = f^T_{\theta}(\tilde{x_i})$ and $h^S = f^S_{\theta}(\tilde{x_i})$ where $h^T, h^S \in \mathbb{R}^D$ and $f^T_{\theta}$ and $f^S_{\theta}$ denote the teacher and student network, respectively. 

We compute the cosine similarity between l2-normalized descriptors from the teacher network and the queue. For the teacher and student, the similarities are:

\vspace*{-10pt}
\begin{eqnarray}
\label{eq:sim}
sim(h^T_i, q^T_j) =  \begin{bmatrix} h^T_i / \lVert h^T_i \rVert_2 \end{bmatrix} \cdot [q_j^T]^{\top} \\
sim(h^{S}_i, q^T_j) =  \begin{bmatrix} h^{S}_i / \lVert h^{S}_i \rVert_2 \end{bmatrix} \cdot [q_j^T]^{\top}
\end{eqnarray}

\noindent where $sim$ represents the cosine similarity, and $q_j^{\top}$ denotes the transposed j-th component in the teacher's queue $Q^T$ to compute cosine similarity. We calculate the probability of the i-th instance with the j-th component in the teacher's queue.

\vspace*{-10pt}
\begin{eqnarray}
\label{eq:prob}
p^T_{i,j} = \frac{exp(sim(h^T_i, q^T_j)/\tau^T)}{\sum_{q^T \sim Q}exp(sim(h^T_i, q^T)/\tau^T)}, \; j \in [1 \dots K]\\
p^S_{i,j} = \frac{exp(sim(h^{S}_i, q^T_j)/\tau^S)}{\sum_{q^T \sim Q}exp(sim(h^{S}_i, q^T)/\tau^S)}, \; j \in [1 \dots K]
\end{eqnarray}

\noindent where $\tau^T$ and $\tau^S$ are the temperature parameters of the teacher and student networks, respectively. K is the size of the teacher's queue. Also, $p^T_{i,j}$ denotes the similarity score between the embeddings $h^T_i$ from the teacher network and the embeddings in the teacher's queue. In the same way,  $p^S_{i,j}$ denotes the similarity score between the embeddings $h^{S}_i$ from the student network and the embeddings in the teacher's queue. The objective of our RDCD is to minimize the KL-divergence between the probabilities over all input instances for the teacher and student networks. The final loss function is determined as follows:

\vspace*{-10pt}
\begin{align}
\label{eq:kl}
\mathcal{L}_{rel} & = \sum_{i}^{N} KL(p^T_i \ || \ p^S_i) \nonumber \\
& = \arg\min_{\theta_S} \sum_i^N\sum_j^K - \frac{\exp(\text{sim}(h^t_i, q^T_j)/\tau^T)}{\sum_{q^T \sim Q}\exp(\text{sim}(h^t_i, q^T)/\tau^T)} \cdot \nonumber \\
& \qquad \log \left( \frac{\exp(\text{sim}(h^s_i, q^T_j)/\tau^S)}{\sum_{q^T \sim Q^T}\exp(\text{sim}(h^s_i, q^T)/\tau^S)} \right)
\end{align}

\subsection{Contrastive Learning for ICD}
The size of the embeddings within the architecture of lightweight networks is generally small, and it is crucial to maintain this size during training while benefiting from knowledge distillation. To ensure this, we attach a linear projector to the end of the student network that is designed to reduce the representation size, resulting in smaller embeddings. We use these final embeddings as our descriptors, which allows for simultaneous training using our relational distillation approach. We follow the MoCo-v2~\cite{MoCov2} training procedure by utilizing a negative instance queue generated by the momentum encoder. We employ the InfoNCE~\cite{infonce} loss to guide contrastive learning as follows:

\begin{eqnarray}
\mathcal{L}_{con} = -\frac{1}{N} \sum_{i=1}^{N} \log \frac{\exp(\text{sim}(z^S_i, z^{S'}_{i}) / \tau)}{\sum_{q^S \sim Q^S} \exp(\text{sim}(z^S_i, q^S) / \tau)}
\end{eqnarray}

For each pair index $i$, $z^S_i$, and $z^{S'}_i$ are the representations of the two augmented views of the same image (i.e., a positive pair). $\text{sim}(z^S_i, z^{S'}_{i})$ is a function measuring the similarity between these descriptors, typically a dot product. $\tau$ is a temperature parameter that scales the similarity scores. The denominator sums the exponentiated similarities of $z^S_i$ with all negative descriptors $q^S$ in all students' queue $Q^S$.

\subsection{The Hard Negative Loss}
In the copy detection scenario, the primary challenges for training arise from the hard negative pairs and the fact that the use of small descriptors renders the environment susceptible to dimensional collapse. To address this, we employ the Hard Negative (HN) loss~\cite{videosimilarity} to finely control the entropy and minimize the hard negative pairs. We also introduce an additional term that ensures that a hard negative pair can push their descriptors apart.

\begin{eqnarray}
    \mathcal{L}_{hn} = -\frac{1}{N} \log \sum_{i=1}^{N} max_{j \in n_{i, j}}(1-S_{i, j})
\end{eqnarray}

\noindent$n_{i, j}$ denotes the row indices of the negative pairs for the i-th row of the similarity matrix. 

The objective function of the student network is a combination of the contrastive loss (con), the Relational Self-supervised Distillation (rel) loss, and the hard negative (HN) loss. The final objective can be expressed as:
\begin{eqnarray}
\label{eq:overall_loss}
\mathcal{L}_{RDCD} = \lambda_{rel}\mathcal{L}_{rel} + \lambda_{con}\mathcal{L}_{con} + \lambda_{hn}\mathcal{L}_{hn}
\end{eqnarray}

\section{Experiments}
\subsection{Dataset}
\noindent\textbf{DISC2021} is a dataset consisting of training, reference and query images, that is used for the Image Similarity Challenge~\cite{disc}. The query image set contains strong auto-augmented and human-augmented images, and distractive images that may not appear in the reference set. The training set is designed to train a model without any labels, and the reference set is used for searching the background as a database.

\noindent\textbf{Copydays}~\cite{jegou2008hamming} is a dataset designed to detect copied content, that consists of 157 original images and 2,000 queries with cropping, jpeg compressing and strong augmentations. We add 10k distractors from YFCC100M~\cite{yfcc}, a common practice~\cite{dino, multigrain} known as CD10K. We evaluate our method based on the mean average precision(mAP) and micro average precision($\mu AP$) for the strongly transformed copies.

\noindent\textbf{NDEC}~\cite{ndec} dataset incorporates HN distractors, making it more advanced than conventional ICD datasets. The basic data is derived from DISC2021 while the HN images are sourced from OpenImage. Similar to DISC2021, the NDEC dataset is organized into training, query, and reference sets. The training data includes 900,000 basic images and adds HN pairs (100,000 x 2 paired images) to enhance the difficulty level of ICD. The test data contains 49,252 query images and 1,000,000 reference images, with 24,252 of the query images being HN.

\subsection{Evaluation Metrics}
To enable a fair comparison with state-of-the-art methods~\cite{sscd, dino}, we use $\mu AP$~\cite{microap} as the evaluation metric. The equation for $\mu AP$ is as follows:

\begin{eqnarray}
\label{eq:general_kd}
\mu AP =   \sum\limits_{i=1}^{N}P(i)\triangle r(i) \in [0, 1]
\end{eqnarray}

\noindent where $P(i)$ is the precision at position $i$ of the sorted list of pairs, $\triangle r(i)$ is the difference in the recall between position i and i-1, and N is the total number of returned pairs for all queries. Any detected pair for a distractor query will decrease the average precision, thus all queries are evaluated together by merging the returned pairs for all queries, sorting them by confidence, and generating a single precision-recall curve. This is different from mAP, also known as macro-AP~\cite{microap}, where the average precision is computed separately per query and then averaged over all queries. $\mu AP$ considers confidence values to capture all queries, while mAP only considers the query in the ground truth. Thus, $\mu AP$ is a more accurate metric in our context because our queries contain the images not contained in the references. The former is more appropriate for ICD, while the latter is typically used in retrieval tasks. We additionally evaluate with $\mu AP_{SN}$. Score normalization is a post-processing technique commonly used in image search to achieve a more uniform distribution. It adjusts the scores based on an estimate of the density around the query and normalizes them using background images that were not involved in training.

\begin{table}[h]
    \caption{Results for the DISC2021 showing different methods, network architectures (RN-50, ViT-B/8, ViT-B/16, ResNeXt-101, EfficientNet-B0), model sizes, and performance metrics (micro Average Precision $\mu$AP and micro Average Precision with score normalization $\mu$AP{\tiny SN}). * means our implementation. $+$ used as teacher model in our experiments.}
    \label{tab:disc}
    \centering
    \resizebox{1\linewidth}{!}
    {
    \begin{tabular}{lcccccc}
    \toprule
    \multicolumn{1}{l}{Method} & Network & $dim$ & $\mu$AP & $\mu$AP{\tiny SN} \\
    \midrule
            Multigrain~\cite{multigrain} & RN-50 & 1500 & 16.5 & 36.5 \\
            Multigrain~\cite{multigrain} & RN-50 & 2048 & 20.5 & 41.7 \\
            DINO$^{+}$~\cite{dino} & ViT-B/8 & 1536 & 32.6 & 54.4 \\
            DINO~\cite{dino} & ViT-B/16 & 1536 & 32.2 & 53.8 \\
            SSCD$^{+}$~\cite{sscd} & RN-50 & 512 & 61.5 & 72.5 \\
            SSCD$^{+}$~\cite{sscd} & ResNeXt-101 & 1024 & 63.7 & 75.3 \\
            \hline
            \textit{using lightweight$^{*}$} \\
            SSCD & EN-B0 & 64 & 38.2 & 48.5 \\
            SSCD & EN-B0 & 128 & 43.5 & 56.2 \\
            SSCD & EN-B0 & 256 & 46.0 & 59.8 \\
            SSCD & EN-B0 & 512 & 43.6 & 61.1 \\
            \hline
            \textit{ours} \\
            RDCD & EN-B0 & 64 & 43.9 & 53.5 \\
            RDCD & EN-B0 & 128 & 50.0 & 61.1 \\
            RDCD & EN-B0 & 256 & \textbf{52.7} & \textbf{65.7}\\
    \bottomrule
    \end{tabular}
    }
    \vspace{-10pt}
\end{table}

\subsection{Training Implementation}
We train all networks using the pretraining parameters from ImageNet-1K. Batch normalization statistics are synchronized across all GPUs. We adhere to the hyperparameters outlined in SimCLR~\cite{simclr}, MoCo-v2~\cite{MoCov2} and SSCD~\cite{sscd}. The batch size is set at N = 256, with a resolution of 224 x 224 for training and 288 x 288 for evaluation. We use a learning rate of 1.0 and a weight decay of $10^{-6}$. We train all models for 100 epochs. We employ a cosine learning rate scheduler with a warmup period of five epochs and use the Adam optimizer. For our knowledge distillation process, we set the temperature at 0.04 for the teacher and 0.07 for the student. All experiments are conducted on 1 NVIDIA A100 80GB GPU. To ensure reproducibility, we maintain consistent seed values throughout all of the experiments. We further discuss the practicality on RDCD in Appendix~\ref{appendix-e}.

\subsection{Results}
Table~\ref{tab:disc} presents the results for the DISC2021 dataset. The first section of the table summarizes the performance of different methods with varying descriptor sizes, while the second section presents the performance of our baseline SSCD~\cite{sscd} strategy using a lightweight network with descriptor sizes of 64, 128, and 256. When training using a lightweight network, we do not use mix-up to ensure a fair comparison of our methods. The third section shows the results of the proposed RDCD approach. When compared with the methods outlined in the first section, our RDCD method achieves competitive performance despite using significantly smaller descriptors and a more compact network size. It is noteworthy that the RDCD method, with a descriptor size of 64, achieves a $\mu$AP{\tiny SN} of 53.5, which is comparable to the DINO method, which utilizes a ViT-B/16 network with a descriptor size of 1536, yielding a $\mu$AP{\tiny SN} of 53.8. This comparison is significant, as the descriptor size of RDCD is 24 times smaller. When evaluated with an equivalent lightweight architecture, RDCD consistently outperforms SSCD across all three dimensions assessed. We use the SSCD model RN-50, which has a $\mu$AP{\tiny SN} of 72.5, as the teacher network to transfer knowledge to the lightweight network. When we apply our methods to EfficientNet-B0, RDCD outperforms SSCD. Remarkably, RDCD with a descriptor size of 128 achieves a $\mu$AP{\tiny SN} of 61.1, which is competitive and matches the performance of SSCD with a descriptor size of 512. This indicates that our method is capable of delivering strong results even with smaller descriptors.

RDCD also outperforms SSCD using the CD10K dataset in Table~\ref{tab:copydays}. Specifically, when using EfficientNet-B0 with a descriptor size of 128, RDCD achieves an mAP of 79.2, which is significantly higher than the best result for SSCD. With a descriptor size of 256, the performance improvement becomes more significant, achieving an mAP of 81.4.

For the NDEC dataset, Table~\ref{tab:ndec} shows that our method shows competitive performance compared to the teacher network SSCD-RN50, which achieves a $\mu$AP{\tiny SN} of 23.3 with a descriptor size of 512. Our proposed RDCD approach, employing EN-B0 network with a descriptor size of 64, 128, and 256 achieves a $\mu$AP{\tiny SN} of 17.3, 19.5, and 21.3, respectively. This demonstrates that across all compact descriptor sizes, RDCD surpasses the performance of the previous SSCD when using the same lightweight architecture.

\begin{table}[t]
    \caption{Results on the CD10K (Copydays+ 10k distractors). We compare different methods, network architectures (RN-50, ViT-B/8, ViT-B/16, ResNeXt-101, EfficientNet-B0), model sizes, and performance metrics (mean Average Precision (mAP) and micro Average Precision ($\mu$AP). * means our implementation.}
    \label{tab:copydays}
    \centering
    \resizebox{0.95\linewidth}{!}
    {
    \begin{tabular}{lcccc}
        \toprule
        Method & Network & $dim$ & mAP & $\mu$AP \\ 
        \midrule
        Multigrain~\cite{multigrain} & RN-50 & 1500 & 82.3 & 77.3 \\
        DINO~\cite{dino} & ViT-B/8 & 1536 & 85.3 & 91.7 \\
        DINO~\cite{dino} & ViT-B/16 & 1536 & 80.7 & 88.7 \\
        SSCD~\cite{sscd} & RN-50 & 512 & 85.0 & 97.9\\
        SSCD~\cite{sscd} & ResNeXt-101 & 1024 & 91.9 & 96.5  \\
        \hline
        SSCD & EN-B0 & 64 & 64.1 & 95.6 \\
        SSCD & EN-B0 & 128 & 71.9 & 97.0 \\
        SSCD & EN-B0 & 256 & 76.6 & 97.4 \\
        SSCD & EN-B0 & 512 & 77.4 & \textbf{97.4} \\
        \hline
        \textit{ours} \\
        RDCD & EN-B0 & 64 & 72.4 & 94.5 \\
        RDCD & EN-B0 & 128 & 79.2 & 96.1 \\
        RDCD & EN-B0 & 256 & \textbf{81.4} & 95.4 \\
        \bottomrule
    \end{tabular}
    }
    \vspace{-10pt}
\end{table}

\begin{table}[h!]
    \caption{Comparison of NDEC results. * means our implementation.}
    \label{tab:ndec}
    \centering
    \resizebox{0.9\linewidth}{!}
    {
    \begin{tabular}{lcccccc}
    \toprule
    \multicolumn{1}{l}{Method} & Network & $dim$ & $\mu$AP & $\mu$AP{\tiny SN} \\
    \midrule
            DINO~\cite{dino} & ViT-B/8 & 1536 & 16.2 & 22.8 \\   
            DINO~\cite{dino} & ViT-B/16 & 1536 & 18.1 & 26.2 \\ 
            SSCD~\cite{sscd} & RN-50 & 512 & 42.4 & 46.6 \\
            \hline
            \textit{using lightweight$^{*}$} \\
            SSCD & EN-B0 & 64 & 28.9 & 32.8 \\
            SSCD & EN-B0 & 128 & 33.0 & 37.5 \\
            SSCD & EN-B0 & 256 & 35.0 & 40.1 \\
            SSCD & EN-B0 & 512 & 34.5 & 40.8 \\  
            \hline
            \textit{ours} \\
            RDCD & EN-B0 & 64 & 31.4 & 34.7 \\
            RDCD & EN-B0 & 128 & 34.9 & 38.9 \\
            RDCD & EN-B0 & 256 & \textbf{37.5} & \textbf{42.5} \\
    \bottomrule
    \end{tabular}
    }
    \vspace{-10pt}
\end{table}

\section{Discussion}
\subsection{Effect of RSD with HN Loss}
\label{5.1}
Our method employs an additional projector to reduce the descriptor size, which creates a network with a two-layer MLP. This configuration can lead to implicit regularization, due to the interaction between the weight matrices across different layers. Implicit regularization typically restricts the ability of a network to learn diverse features, leading to dimensional collapse with contrastive SSL~\cite{directclr}. However, when applying HN loss together with RSD, we show that dimensional collapse does not occur even with the use of an MLP, thus reducing the impact of implicit regularization. 

\begin{figure}[t]
\centering
\includegraphics[width=1.0\linewidth]{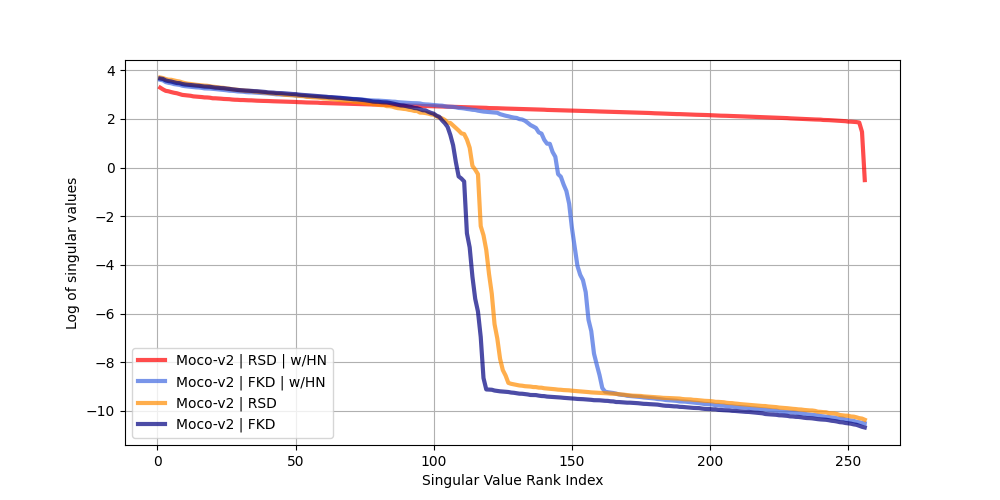}
\caption{Log of singular values for a descriptor size of 256, with and without HN loss.}
\label{fig:collapse}
\vspace{-10pt}
\end{figure}

\begin{figure}[t]
  \centering
  \begin{subfigure}{.33\columnwidth}
    \centering
    \includegraphics[width=\linewidth]{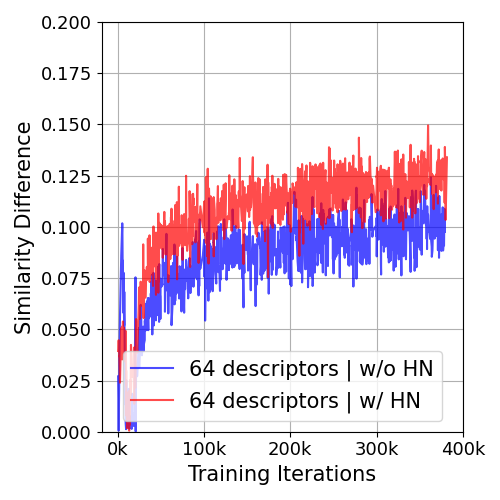}
    \caption{64 descriptors}
    \label{fig:sub1}
  \end{subfigure}%
  \hfill
  \begin{subfigure}{.33\columnwidth}
    \centering
    \includegraphics[width=\linewidth]{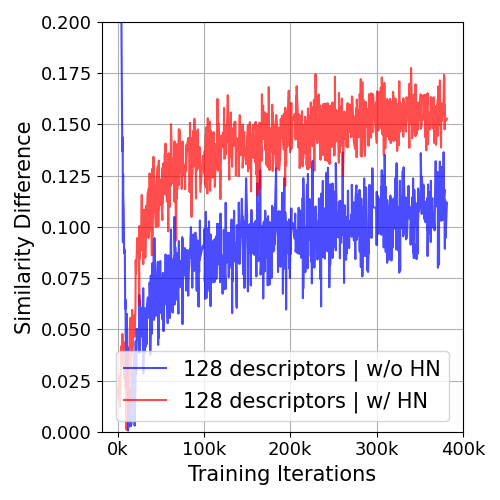}
    \caption{128 descriptors}
    \label{fig:sub2}
  \end{subfigure}
  \begin{subfigure}{.33\columnwidth}
    \centering
    \includegraphics[width=\linewidth]{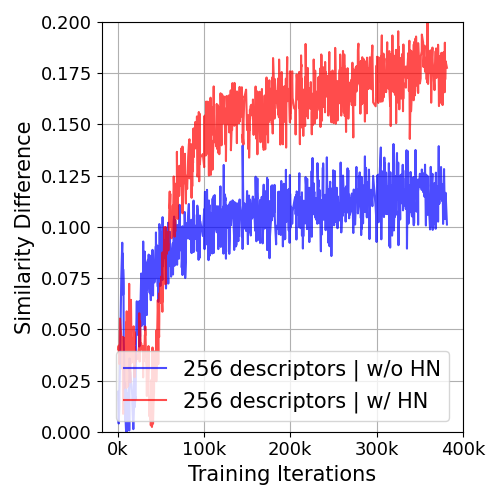}
    \caption{256 descriptors}
    \label{fig:sub2}
  \end{subfigure}
  \caption{Comparison of the difference in similarity between positives and nearest negatives with and without the use of HN Loss.}
  \label{fig:diff}
  \vspace{-10pt}
\end{figure}

To verify this, we calculate the singular values of the final descriptor of a model with a descriptor size of 256 under four conditions: with the use of RSD, FKD and with and without the use of HN loss. We compute the descriptors for 50k queries from the DISC2021 dataset, calculate their covariance, and perform singular value decomposition on the covariance matrix. Subsequently, we extract the singular values and apply a logarithmic operation to them. As illustrated in Figure~\ref{fig:collapse}, RDCD has a full rank, while the absence of HN loss results in dimensional collapse. It can also be observed that applying HN loss with FKD does not have a significant effect. We further visualize the difference between positive similarity and hard negative similarity across multiple dimensions experimented in MoCo-v2 $|$ RSD environment, comparing scenarios with and without the application of HN loss. We analyze a sample of 5,000 distinct queries extracted from the DISC2021 dataset, which are separate from the queries used in the standard evaluation process for generality, to calculate and assess the differences in the similarity between positive and hard negative samples. As Figure~\ref{fig:diff} shows, across all descriptor sizes, the application of HN loss resulted in a more pronounced disparity compared to when it was omitted. This observation implies that HN loss effectively enhances the separation between positive and hard negative samples. Furthermore, we noted that as the dimensionality increases, the difference between positive and hard negative similarity also enlarges. We conjecture that this trend may contribute to improved performance in higher-dimensional descriptors.

\subsection{Effect of Hard Negative Loss on Uniformity}
\begin{figure}[t]
    \centering
    \includegraphics[width=1.0\linewidth]{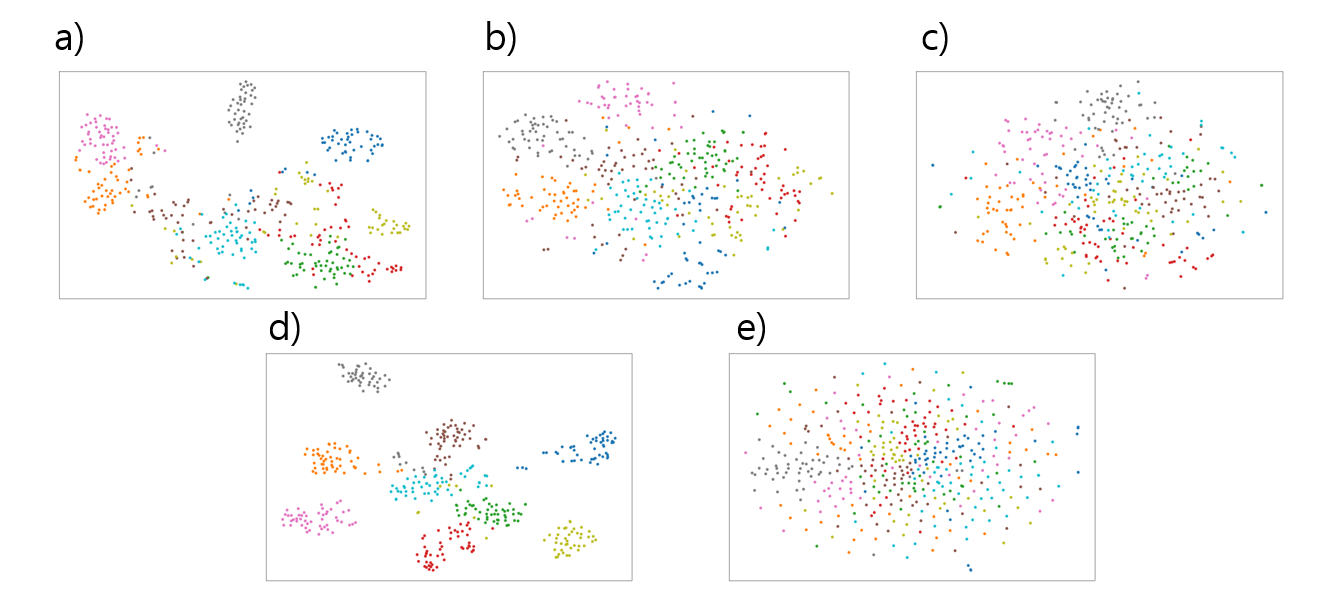}
    \caption{T-SNE visualization of descriptors from different model configurations: a) DINO baseline, b) DINO-MoCo-v2 w/ HN loss, c) SSCD-MoCo-v2 w/ HN loss. d) DINO-MoCo-v2 w/o HN loss, e) SSCD-MoCo-v2 w/o HN loss. In a),b) and c), the final features are used as descriptors whereas in d) and e), intermediate features are used. We randomly select 9 classes and visualize 50 images from each class from ImageNet.}
    \label{fig:uniform}
\vspace{-10pt}
\end{figure}

A uniform distribution of descriptors represents semantically similar images far apart while placing copies of the same image close together. Both DINO~\cite{dino} and SSCD~\cite{sscd} are used as teacher models; however, only SSCD uses entropy regularizer to achieve a uniform distribution. 
In general, features used during evaluation are called descriptors. We demonstrate that applying HN loss to the final feature used as a descriptor results in a more uniform distribution. This effect occurs regardless of whether the teacher was trained with or without HN loss. Figure~\ref{fig:uniform} visually supports this observation. b) shows the distribution of a student model distilled from a non-uniform teacher and trained with HN loss, while c) displays the distribution of a student model distilled from a uniform teacher, also trained with HN loss. Both distributions are notably more uniform compared to the distribution of the DINO teacher in a). 
We further analyze the intermediate feature, which is directly distilled by the teacher model, and show that when using RSD, the uniformity of the teacher’s embedding is transferred to the student’s embedding. d) shows the result of distillation from a non-uniform teacher DINO, while e) shows the result of distillation from a uniform teacher SSCD. Consequently, if the teacher’s embedding has a uniform distribution in the embedding space, applying HN loss to the final feature may yield limited additional uniformity. Additional experiments with intermediate features are illustrated in Appendix~\ref{appendix-b}.

\subsection{Ablation Study}
\textbf{Effectiveness of Each Loss Term.} To further assess the effectiveness of RDCD, we investigate the impact of different loss components on its performance, specifically, RSD loss and HN loss as shown in Table~\ref{tab:ablation1}. In this section, we present an additional metric, the rank preserving ratio (RPR), which is the rank divided by the descriptor dimension. A low RPR indicates dimensional collapse, and RPR value of 1 indicates that the descriptor makes full use of all dimensions. All models are trained on the DISC2021 dataset with 100 epochs and SSCD-RN50-w/mixup is employed as the teacher model. Our result shows that with the addition of RSD loss and HN loss, the performance improves substantially, which indicates that RSD and HN loss can benefit the performance of RDCD.

\begin{table}[h]
  \caption{Ablation study on the loss terms in RDCD. Descriptor sizes of 64, 128, and 256 were examined, and a consistent MLP configuration was maintained across all experiments for a fair comparison.}
  \centering
  \label{tab:ablation1}
  \resizebox{1.0\linewidth}{!}{
    \begin{tabular}{cccccccc}
    \toprule
    \multicolumn{3}{c}{Loss} & \multicolumn{2}{c}{\multirow{2}{*}{MLP-d}} & \multirow{2}{*}{$\mu$AP} & \multirow{2}{*}{$\mu$AP{\tiny SN}}  \\
    $\mathcal{L}_{con}$ & $\mathcal{L}_{rel}$ & $\mathcal{L}_{hn}$ & & &  & \\ 
    \midrule
    \multicolumn{2}{l}{\textit{MoCo-v2}} \\
    \checkmark & & & 1280/512/64 & & 10.9 & 14.4  \\
    \checkmark & & & 1280/512/128 & & 11.9 & 17.2  \\
    \checkmark & & & 1280/512/256 & & 16.5 & 26.4  \\
    \multicolumn{2}{l}{\textit{Effectiveness of RSD loss}} \\
    \checkmark & \checkmark & & 1280/512/64 & & 39.9 & 52.1  \\
    \checkmark & \checkmark & & 1280/512/128 & & 47.1 & 60.7  \\
    \checkmark & \checkmark & & 1280/512/256 & & 47.2 & 61.0 \\
    \multicolumn{2}{l}{ \textit{Effectiveness of HN loss}} \\   
    \checkmark & \checkmark & \checkmark & 1280/512/64 & & 43.9 & 53.5 \\
    \checkmark & \checkmark & \checkmark & 1280/512/128 & & 50.0 & 61.1  \\
    \checkmark & \checkmark & \checkmark & 1280/512/256 & & \textbf{52.9} & \textbf{65.7} \\
    \bottomrule
    \end{tabular}
  }
\end{table}

\noindent\textbf{Influence of $\lambda_{rel}$ and $\lambda_{hn}$.}
$\lambda_{rel}$ and $\lambda_{hn}$ are the weights of RSD loss and HN loss, respectively. First, we analyze the influence of contrastive loss and RSD loss, as shown in Table~\ref{tab:ablation2}. As shown in the third row, a contrastive loss of 1 and RSD loss of 10 achieves the best performance, which indicates that knowledge distillation is essential for the model with better performance. The HN loss is also varied with the contrastive loss and RSD loss fixed at 1 and 10, respectively, showing that the best performance occurs when the HN loss is 5. However, for larger values, the performance drops to 0, (i.e. collapses) because the network focuses more on HN loss rather than contrastive and distillation loss. We further calculate RPR of each experiment. In the first section, where only contrastive loss and RSD loss are employed, we observe that RPR gradually increases incrementally with an increase in $\lambda_{rel}$. This suggests that the student model partially alleviates the dimensional collapse which is guided by a teacher model that has been trained with an entropy regularizer. Subsequently, in the second section, where HN loss is incorporated, the student model finally utilizes the full dimensions of the descriptors.

\begin{table}[h]
    \caption{Ablation study to evaluate the impact of the loss ratio. All experiments are performed with a descriptor size of 256. }
    \label{tab:ablation2}
    \centering
    \resizebox{0.8\linewidth}{!}
    {
    \begin{tabular}{cccccc}
        \hline
        $\lambda_{con}$ & $\lambda_{rel}$ & $\lambda_{hn}$ & RPR & $\mu$AP & $\mu$AP{\tiny SN} \\
        \hline
        10 & 1 & - & 0.27 & 17.0 & 32.5 \\
        1 & 1 & - & 0.40 & 33.4 & 52.4 \\
        1 & 10 & - & 0.55 & 47.2 & 61.0 \\
        \hline
        1 & 10 & 1 & 0.62 & 47.3 & 61.4 \\
        1 & 10 & 3 & 0.98 & 47.1 & 62.1 \\
        1 & 10 & 5 & 1.0 & 52.9 & \bf{65.7} \\
        1 & 10 & above 5 & - & \textit{collapse} & \textit{collapse} \\
        \hline
    \end{tabular}
    }
\end{table}

We next conduct a step-by-step experiment with DINO as the teacher model, as shown in Table~\ref{tab:dino}. Our approach reduces the descriptor size by a factor of 12 but produces a performance comparable to that of the teacher model. As illustrated in \ref{5.1}, dimensional collapse occurs when HN loss is not employed. Applying HN loss thus successfully improves performance and prevents dimensional collapse.

\begin{table}[h]
    \caption{Ablation study of using DINO as teacher model.}
    \label{tab:dino}
    \centering
    \resizebox{1\linewidth}{!}
    {
    \begin{tabular}{lccccc}
        \toprule
        Method & Network & $dim$ & RPR & $\mu$AP & $\mu$AP{\tiny SN} \\
        \midrule
        DINO~\cite{dino} (Teacher) & ViT-B/8 & 1536 & 0.99 & 32.6 & 54.4  \\
        RSD & EN-B0 & 1536 & 0.88 & 10.0 & 35.2 \\
        RSD $|$ MoCo-v2  & EN-B0 & 128 & 0.57 & 20.7 & 40.0  \\
        RSD $|$ MoCo-v2 $|$ HN (RDCD) & EN-B0 & 128 & 1.0 & {\bf 32.1} & {\bf 52.1}  \\
        \bottomrule
    \end{tabular}
    }
\end{table}

\section{Conclusion}
In this paper, we present RDCD, a novel method for training lightweight networks with compact descriptors for image copy detection in a self-supervised manner. We demonstrate in a series of experiments that RDCD, which combines RSD with HN loss effectively prevents dimensional collapse in lightweight architectures, and achieves a competitive performance across various benchmarks. We believe that this approach offers significant advantages in search speed and scalability for multimedia applications.

\section*{Acknowledgements}
This work was supported by Institute of Information \& communications Technology Planning \& Evaluation (IITP) grant funded by the Korea government(MSIT) (RS-2023-00224740, Development of technology to prevent and track the distribution of illegally filmed content).

\newpage
{\small
\bibliographystyle{ieee_fullname}
\bibliography{egbib}
}

\clearpage
\section*{Appendix}
\appendix

\section{Effect of Different Architectures}
\subsection{Different Student}
To show the generality of RDCD, we adopt one lightweight model each from the most commonly used architectures in computer vision, CNN and ViT, conducting experiments with MobileNet-V3 and FastViT-T12. Our results are shown in Table~\ref{tab:disc-appen}, Table~\ref{tab:copydays-appen}, Table~\ref{tab:ndec-appen}. Specifically, using FastViT under the same conditions resulted in a $\mu$AP of 67.4, demonstrating the research potential for ViT student.

\begin{table}[h]
    \caption{Performance of DISC2021 with different student  architectures. * means our implementation.}
    \label{tab:disc-appen}
    \centering
    \resizebox{1\linewidth}{!}
    {
    \begin{tabular}{lcccc}
        \toprule
        Method & Network & $dim$ & $\mu$AP & $\mu$AP{\tiny SN} \\ 
        \midrule
        \textit{using lightweight $^{*}$} \\
        SSCD & Mob-V3 & 128 & 45.5 & 56.2 \\
        SSCD & FastViT-T12 & 128 & 42.5 & 59.7 \\
        SSCD & Mob-V3 & 256 & 47.3 & 60.1 \\
        SSCD & FastViT-T12 & 256 & 43.1 & 59.5 \\
        \hline
        \textit{ours } \\
        RDCD & Mob-V3 & 128 & 50.6 & 60.8 \\
        RDCD & FastViT-T12 & 128 & 48.5 & 61.1 \\
        RDCD & Mob-V3 & 256 & 53.9 & 65.6 \\
        RDCD & FastViT-T12 & 256 & 56.4 & \textbf{67.4} \\
        \bottomrule
    \end{tabular}
    }
    \vspace{-10pt}
\end{table}

\begin{table}[h]
    \caption{Performance of CD10K(Copydays+ 10k distractors) with different student architectures. * means our implementation.}
    \label{tab:copydays-appen}
    \centering
    \resizebox{1\linewidth}{!}
    {
    \begin{tabular}{lcccc}
        \toprule
        Method & Network & $dim$ & mAP & $\mu$AP \\ 
        \midrule
        Multigrain & RN-50 & 1500 & 82.3 & 77.3 \\
        DINO & ViT-B/8 & 1536 & 85.3 & 91.7 \\
        DINO & ViT-B/16 & 1536 & 80.7 & 88.7 \\
        SSCD & RN-50 & 512 & 85.0 & 97.9\\
        SSCD & ResNext-101 & 1024 & 91.9 & 96.5  \\
        \hline
        \textit{using lightweight $^{*}$} \\
        SSCD & Mob-V3 & 128 & 68.9 & 97.1 \\
        SSCD & FastViT-T12 & 128 & 60.9 & 87.0 \\
        SSCD & Mob-V3 & 256 & 75.4 & 97.4 \\
        SSCD & FastViT-T12 & 256 & 74.4 & 92.2 \\
        \hline
        \textit{ours } \\
        RDCD & Mob-V3 & 128 & 81.3 & 96.5 \\
        RDCD & FastViT-T12 & 128 & 77.4 & 95.3 \\
        RDCD & Mob-V3 & 256 & 83.5 & \textbf{97.6} \\
        RDCD & FastViT-T12 & 256 & 79.3 & 97.0 \\
        \bottomrule
    \end{tabular}
    }
    \vspace{-10pt}
\end{table}

\begin{table}[h!]
    \caption{Performance of NDEC with different student architectures. * means our implementation.}
    \label{tab:ndec-appen}
    \centering
    \resizebox{1\linewidth}{!}
    {
    \begin{tabular}{lcccccc}
    \toprule
    \multicolumn{1}{c}{Method} & Network & $dim$ & $\mu$AP & $\mu$AP{\tiny SN} \\
    \midrule
            DINO & ViT-B/8 & 1536 & 16.2 & 22.8 \\   
            DINO & ViT-B/16 & 1536 & 18.1 & 26.2 \\ 
            SSCD & RN-50 & 512 & 42.4 & 46.6 \\
            \hline
            \textit{using lightweight $^{*}$} \\
            SSCD & Mob-V3 & 128 & 34.6 & 38.9 \\
            SSCD & FastViT-T12 & 128 & 33.5 & 38.6 \\
            SSCD & Mob-V3 & 256 & 36.8 & 41.5 \\
            SSCD & FastViT-T12 & 256 & 33.3 & 38.3 \\  
            \hline
            \textit{ours} \\
            RDCD & Mob-V3 & 128 & 35.5 & 39.3 \\
            RDCD & FastViT-T12 & 128 & 35.1 & 38.7 \\
            RDCD & Mob-V3 & 256 & 37.9 & 42.6 \\
            RDCD & FastViT-T12 & 256 & 39.1 & \textbf{43.2} \\
    \bottomrule
    \end{tabular}
    \vspace{-10pt}
    }
\end{table}

\subsection{Different Teacher}
We demonstrate the effectiveness of our proposed approach in scenarios where the teacher networks belong to different architectural families (ViT or CNN), with student models also selected from these two distinct architectures. For the ViT, we use DINO as the teacher, and for the CNNs, we use SSCD-RN50 and SSCD-ResNeXt-101. All models are trained on SSCD-RN50 hyperparameter settings.
With DINO as the teacher model, we follow the ICD settings presented in~\cite{dino} for the DINO baseline. Consequently, we distill a descriptor size of 1536 from the teacher (concatenation of 768 class tokens and 768 GeM pooled patch tokens). In the case of ResNeXt-101 teacher model, we distill the final target descriptor with a size of 1024. Table~\ref{tab:architectures} shows that our method is highly effective for the different architectures, highlighting its adaptability and robustness. 

\begin{table}[h]
    \caption{DISC2021 evaluation with different architectures. All experiments are performed with a descriptor size of 128.}
    \label{tab:architectures}
    \centering
    \resizebox{1\linewidth}{!}
    {
    \begin{tabular}{cc|cc|cc|cc}
        \hline
        \multicolumn{2}{c|}{{\multirow{2}{*}{\diagbox{T}{S}}}} & \multicolumn{2}{c|}{EN-B0} & \multicolumn{2}{c|}{FastViT-T12} & \multicolumn{2}{c}{Mob-V3} \\  
        \cline{3-8}
        & & $\mu$AP & $\mu$AP{\tiny SN} & $\mu$AP & $\mu$AP{\tiny SN} & $\mu$AP & $\mu$AP{\tiny SN} \\
        \hline
        \multicolumn{2}{c|}{DINO-ViT-B/8} & 32.1 & 52.1 & 28.1 & 52.2 & 39.6 & 54.7 \\
        \multicolumn{2}{c|}{SSCD-RN-50} & 50.0 & 61.1 & 48.5 & 61.1 & 47.3 & 58.8 \\
        \multicolumn{2}{c|}{SSCD-ResNeXt-101} & 44.9 & 59.9 & 42.8 & 57.4 & 41.1 & 56.6 \\ 
        \hline
    \end{tabular}
    }
    \vspace{-10pt}
\end{table}

\begin{table*}[h!]
    \caption{The following table demonstrates that when dimension reduction is applied to the descriptor through PCA whitening, our approach outperforms the existing baseline in terms of $\mu$AP performance. We do not apply Mixup. * means our implementation. }
    \label{tab:pca}
    \centering
    \resizebox{0.8\linewidth}{!}
    {
    \begin{tabular}{cccccccc}
        \hline
        Method & Network & $dim$ & $\mu$AP & $\mu$AP{\tiny SN} & $\mu$AP{\tiny SN}256 &  $\mu$AP{\tiny SN}128 & $\mu$AP{\tiny SN}64\\
        \midrule
        DINO & ViT-B/8 & 1536 & 32.6 & 54.4 & 47.7 & 42.3 & 32.7 \\
        DINO & ViT-B/16 & 1536 & 32.2 & 53.8 & 47.4 & 42.0 & 32.0 \\
        SSCD & RN-50 & 512 & 43.6 & 72.5 & 66.0 & 56.3 & 38.3 \\
        SSCD & ResNext-101 & 1024 & 63.7 & 75.3 & 65.5 & 54.0 & 34.8 \\
        \hline
        \textit{using lightweight $^{*}$} \\
        SSCD & EN-B0 & 64 & 38.2 & 48.5 & 48.5 & 48.5 & 48.5  \\
        SSCD & EN-B0 & 128 & 43.5 & 56.2 & 56.2 & 56.2 & 44.4 \\
        SSCD & EN-B0 & 256 & 46.0 & 59.8 & 59.8 & 53.8 & 42.3 \\
        SSCD & EN-B0 & 512 & 43.6 & 61.1 & 58.0 & 52.0 & 41.9\\
        \hline
        \textit{Ours} \\
        RDCD & EN-B0 & 64 & 43.9 & 53.5 & 53.5 & 53.5 & \textbf{53.5} \\
        RDCD & EN-B0 & 128 & 50.0 & 61.1 & 61.1 & 61.1 & 53.1\\
        RDCD & EN-B0 & 256 & 52.7 & \textbf{65.7} & \textbf{65.7} & \textbf{61.5} & 52.2 \\
        \hline
    \end{tabular}
    }
    \vspace{-10pt}
\end{table*}

\begin{table}[h]
\caption{Comparison of intermediate descriptor. We do not use HN loss for SSCD method as they use Koleo regularizer.}
\label{tab:13}
\centering
\resizebox{1.0\linewidth}{!}{
\begin{tabular}{lcccccc}
\textbf{\textit{128 projection}} \\
\toprule
\multicolumn{3}{l}{Hard Negative Loss} & \multicolumn{2}{c}{No} & \multicolumn{2}{c}{Yes} \\
Method & RSD & FKD & $\mu$AP & $\mu$AP{\tiny SN} & $\mu$AP & $\mu$AP{\tiny SN} \\
\midrule
SSCD(RN50) w/o mixup         &  &  & 60.4 & 71.1 & - & - \\
SSCD(EN-B0) w/o mixup        &  &  & 43.6 & 61.1 & - & - \\
SimCLR       & & \checkmark & 56.1 & 68.0 & 55.2 & 67.6 \\
             & \checkmark & & 56.3 & 69.7 & 56.3 & 69.7 \\
             & \checkmark & \checkmark & 56.6 & 69.5 & 56.7 & 69.6 \\
MoCo-v2       & & \checkmark & 51.2 & 67.3 & 54.4 & 67.8 \\
             & \checkmark & & 55.1 & 69.1 & 55.5 & 69.1 \\
             & \checkmark & \checkmark & 55.2 & 69.1 & 55.8 & 68.9 \\
\end{tabular}
}

\centering
\resizebox{1.0\linewidth}{!}{
\begin{tabular}{lcccccc}
\textbf{\textit{256 projection}} \\
\toprule
\multicolumn{3}{l}{Hard Negative Loss} & \multicolumn{2}{c}{No} & \multicolumn{2}{c}{Yes} \\
Method & RSD & FKD & $\mu$AP & $\mu$AP{\tiny SN} & $\mu$AP & $\mu$AP{\tiny SN} \\
\midrule
SSCD(RN50) w/o mixup         &  &  & 60.4 & 71.1 & - & - \\
SSCD(EN-B0) w/o mixup        &  &  & 43.6 & 61.1 & - & - \\
SimCLR       & & \checkmark & 56.2 & 68.1 & 55.5 & 67.1 \\
             & \checkmark & & 56.3 & 69.5 & 56.4 & 69.5 \\
             & \checkmark & \checkmark & 55.7 & 67.9 & 54.1 & 68.5 \\
MoCo-v2       & & \checkmark & 53.3 & 67.4 & 54.3 & 67.7 \\
             & \checkmark & & 53.7 & 68.2 & 55.4 & 68.9 \\
             & \checkmark & \checkmark & 55.8 & 69.2 & 56.1 & 69.0  \\
\bottomrule
\end{tabular}
\vspace{-10pt}
}

\end{table}

\section{Intermediate Descriptor}
\label{appendix-b}
In this study, we evaluate the performance of DISC2021 using an intermediate descriptor designed to align the descriptor size with that of the teacher model. We assess a 512-dimensional descriptor both with and without the application of HN loss, utilizing the EfficientNet-B0 network architecture. Our results (Table~\ref{tab:13}) show that the proposed RDCD method consistently outperforms SSCD across two projection sizes, 128 and 256. Furthermore, our analysis reveals that the performance gap between using and not using HN loss is negligible, and the intermediate descriptor fully exploits its 512-dimensional capacity without HN loss. This finding suggests that the student model is effectively trained under the guidance of a teacher model equipped with an entropy regularizer.

\section{PCA}
The dimensionality of the descriptor plays a crucial role in optimizing the trade-off between matching time and accuracy during the matching phase of ICD. Previous approaches have employed PCA and PCA whitening to create compact descriptors. Following~\cite{sscd}The post-processing technique of PCA whitening is known for its efficacy in making the descriptor distribution more uniform. Our RDCD method, trained via deep learning, exhibits superior performance compared to the baseline, which utilizes large-sized descriptors that have undergone PCA whitening. These results are detailed in Table ~\ref{tab:pca}.

\begin{table}[h]
\caption{Comparison of other distillation. All methods are trained with EfficientNet-B0 network. We do not use Hard Negative loss for SSCD method as they use Koleo regularizer.}
\label{tab:projection}
\centering
\resizebox{1.0\linewidth}{!}{
\begin{tabular}{lcccccc}
\textbf{\textit{128 descriptors}} \\
\toprule
\multicolumn{3}{l}{Hard Negative Loss} & \multicolumn{2}{c}{No} & \multicolumn{2}{c}{Yes} \\
Method & RSD & FKD & $\mu$AP & $\mu$AP{\tiny SN} & $\mu$AP & $\mu$AP{\tiny SN} \\
\midrule
SSCD         &  &  & 43.5 & 56.2 & - & - \\
SimCLR       & & \checkmark & 46.2 & 58.7 & 48.4 & 59.9 \\
             & \checkmark & & 47.4 & 59.6 & 49.6 & \underline{61.4} \\
             & \checkmark & \checkmark & 47.5 & 59.6 & 50.8 & \textbf{61.5} \\
MoCo-v2       & & \checkmark & 40.3 & 57.5 & 44.8 & 59.1 \\
             & \checkmark & & 47.1 & 60.7 & 50.0 & \textbf{\underline{61.1}} \\
             & \checkmark & \checkmark & 46.2 & 59.3 & 49.6 & 61.0 \\
\end{tabular}
}
\centering
\resizebox{1.0\linewidth}{!}{
\begin{tabular}{lcccccc}
\textbf{\textit{256 descriptors}} \\
\toprule
\multicolumn{3}{l}{Hard Negative Loss} & \multicolumn{2}{c}{No} & \multicolumn{2}{c}{Yes} \\
Method & RSD & FKD & $\mu$AP & $\mu$AP{\tiny SN} & $\mu$AP & $\mu$AP{\tiny SN} \\
\midrule
SSCD         &  &  & 46.0 & 59.8 & - & - \\
SimCLR       & & \checkmark & 51.7 & 64.0 & 51.4 & 64.5 \\
             & \checkmark & & 52.2 & 65.6 & 52.7 & \textbf{\underline{65.8}} \\
             & \checkmark & \checkmark & 52.4 & 65.5 & 49.9 & 64.6 \\
MoCo-v2       & & \checkmark & 41.3 & 56.6 & 46.1 & 60.1 \\
             & \checkmark & & 47.2 & 61.0 & 52.9 & \underline{65.7} \\
             & \checkmark & \checkmark & 47.3 & 60.8 & 53.2 & \textbf{65.9}  \\
\bottomrule
\end{tabular}
}
\end{table}

\section{Comparison with Other Forms of Distillation}
To further demonstrate the efficacy of RSD, we conduct a comparative analysis with FKD by 1) employing FKD alone, 2) employing RSD alone, and 3) integrating both FKD and RSD in Table~\ref{tab:projection}. We also employ SimCLR-style contrastive learning. When FKD and RSD are combined, the best performance is observed in several cases, suggesting that combining both distillation methods can further improve performance. Nevertheless, RSD consistently outperforms FKD in all cases, with RSD alone achieving performance on par with the combined use of both distillation methods. This indicates that RSD is sufficiently robust, and FKD does not significantly alter the outcome. It confirms that RSD can operate alone to enhance the performance of compact descriptors.

\begin{table}[t]
    \small
    \centering
    \caption{Ablation study on the computational cost during training. All experiments are conducted on a single A100 GPU using FP32 precision with EFF-B0 architecture and a descriptor size of 256. All metrics are calculated with a batch size of 256 except for FLOPs, which we measured with a batch size of 1.}
    \label{tab:train}
    \resizebox{1.0\linewidth}{!}{
    \begin{tabular}{lccccccc}
    \toprule
     Method & Contrastive Learning & Distillation & \#Params & FLOPs & Time per Epoch & VRAM \\
           &  &  & (M) & (G) & (min) & (G) \\
    \midrule
    SSCD & SimCLR & X & 4.3 & 0.828 & 58.46 & 49.7 \\ 
     - & - & O & 29.3 & 8.732 & 40.06 & 28.1 \\
     - & SimCLR & O & 29.4 & 9.16 & 74.34 & 54.9\\
    \textbf{RDCD} & MoCo-v2 & O & 34.2 & 9.9 & 51.25 & 29.8 \\
    \bottomrule
  \end{tabular} 
  }
\end{table}

\begin{table}[t]
    \small
    \centering
    \caption{Comparison of computational cost during evaluation, highlighting our method's high efficiency. We evaluate images with 288x288 size. FLOPs is calculated with a batch size of 1.}
    \label{tab:eval}
    \resizebox{1.0\linewidth}{!}{
    \begin{tabular}{lcccccccc}
    \toprule
     Method & Architecture & \#Params & FLOPs & Throughput & Search Time & $dim$ & $\mu$AP \\
           &  & (M) & (G) & (image/sec) & & & \\
    \midrule
    DINO & ViT-B/8 & 85.204 & 133.699 & 87.76 & 37.5 & 1536 & 54.4 \\ 
    SSCD & RN-50 & 24.6 & 8.265 & 903.58 & 15.518 & 512 & 72.5\\
    \textbf{RDCD} & EN-B0 & \textbf{4.8} & \textbf{0.829} & \textbf{1087.75} & \textbf{10.121} & 256 & 65.7 \\
    \bottomrule
  \end{tabular} 
  }
  \vspace{-10pt}
\end{table}

\section{Practicality of RDCD}
\label{appendix-e}
We perform additional analysis of practicality on RDCD. To highlight the computational cost of each component, in Table~\ref{tab:train} we conduct ablation experiments by removing contrastive learning and distillation. SSCD uses SimCLR without distillation. We found that our method requires 29.9 million more parameters than SSCD for training, primarily due to the large teacher network, which accounts for 24.0M parameters. Additionally, our method consumes a total of 9.9G of FLOPs: 8.3G for distillation and 1.6G for contrastive learning. Nevertheless, SSCD's use of SimCLR leads to a proportional increase in VRAM usage with larger batch sizes. In our RDCD, we adopt MoCo-v2 which stores negative features in a queue, significantly reducing VRAM usage. By leveraging MoCo-v2, our method achieves 0.6 times lower VRAM usage compared to SSCD and also reduces training time. Furthermore, our final performance surpasses SSCD by 5.9.

Our RDCD incurs higher computational cost during training but shows significant efficiency during evaluation. Specifically, RDCD requires only 4.8M parameters during evaluation, which includes an 0.5M additional parameters from the EFF-B0 architecture. In copy detection, the evaluation involves two stages: model inference and database search. High-speed search is crucial when retrieving from a large-scale image database in real-world applications. We evaluate search time using the DISC2021 dataset, which consists of 1M database images and 50k query images. As shown in  Table~\ref{tab:eval}, RDCD achieves superior efficiency across all metrics compared to previous methods. DINO and SSCD require 133.699G, 8.265G FLOPs, respectively, due to their large architectures, while RDCD operates with 161 times lower FLOPs than DINO and 9.9 times lower FLOPs than SSCD during model inference.  

\section{Qualitative Examples}
\begin{figure}[h]
    \centering
    \includegraphics[width=0.9\linewidth]{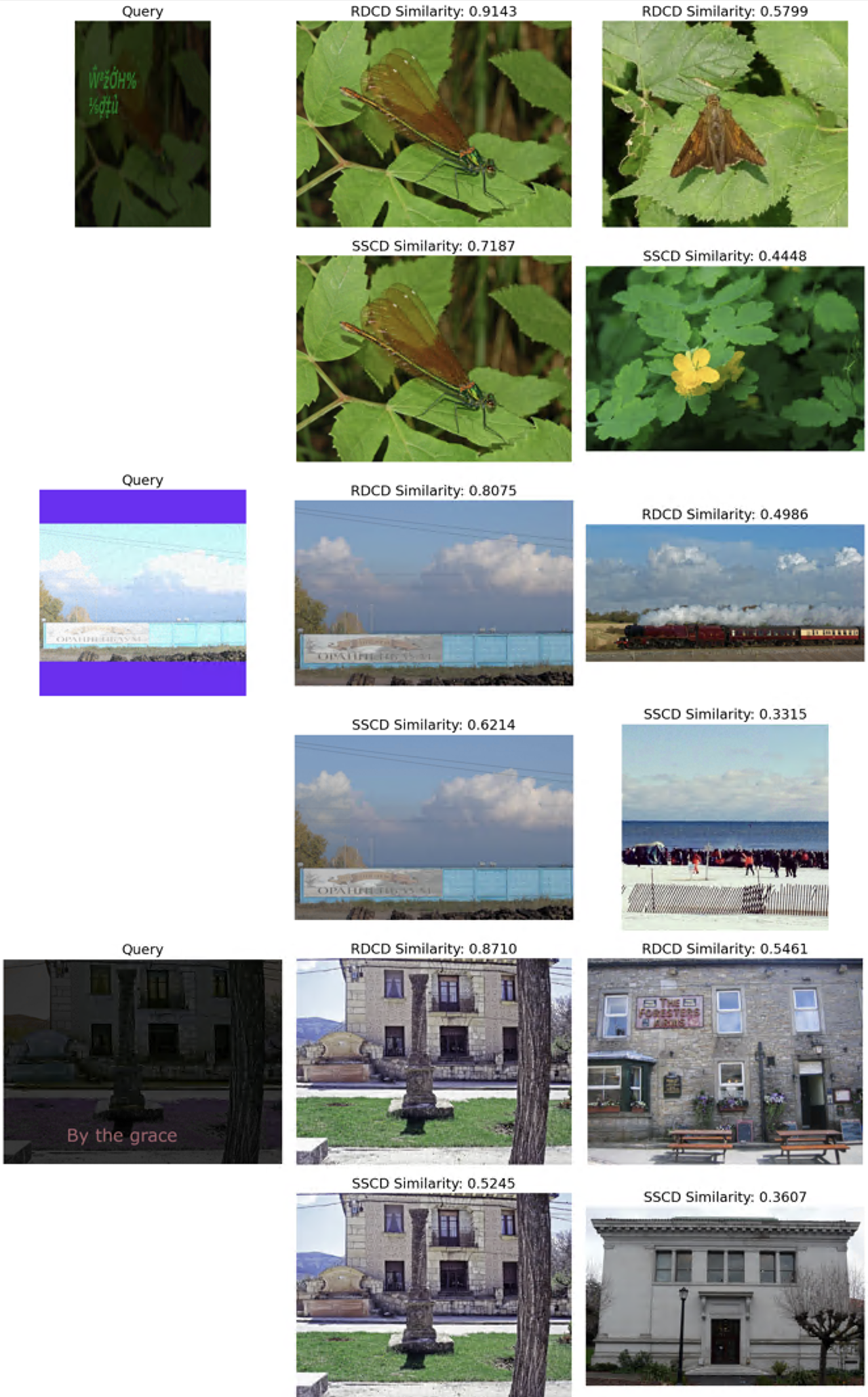}
    \caption{Example of top-2 retrieval results from the DISC2021 dataset.}
    \label{fig:qualitative}
\end{figure}

In Figure~\ref{fig:qualitative}, we present the queries along with the top-2 results retrieved by both the RDCD and SSCD methods for cases where both methods accurately identify the ground truth. However, it is noteworthy that RDCD consistently retrieves a smaller similarity score difference between the ground truth and the hardest negative sample, compared to SSCD. For our matching evaluation, we employ a descriptor of size 256, utilizing the EN-B0 network architecture.

\end{document}